\documentclass[10pt,twocolumn,letterpaper]{article}

\usepackage{times}
\usepackage{epsfig}
\usepackage{graphicx}
\usepackage{amsmath}
\usepackage{amssymb}
\usepackage{tabularx}
\usepackage{booktabs}
\usepackage{multirow}
\usepackage{cancel}
\usepackage{amssymb}
\usepackage{subfig}
\usepackage{amsthm}
\usepackage{footnote}
\usepackage{eso-pic}
\usepackage{xspace}
\usepackage{mathtools}

\newcommand{\bt}{\textbf}
\DeclareMathOperator*{\argmax}{arg\; max}
\DeclareMathOperator*{\argmin}{arg\; min}

\newcommand{\mb}[1]{\mathbf{#1}}

\newcommand{\svmmid}{MidRank\xspace}

\newcommand{\lrank}{learning to rank\xspace}

\newtheorem{prop}{Proposition}

\makeatletter
\DeclareRobustCommand\onedot{\futurelet\@let@token\@onedot}
\def\@onedot{\ifx\@let@token.\else.\null\fi\xspace}

\def\eg{\emph{e.g}\onedot} 
\def\ie{\emph{i.e}\onedot}

\def\etal{\emph{et al}\onedot}
\makeatother

\usepackage[breaklinks=true,bookmarks=false]{hyperref}

\begin{document}

\title{MidRank: Learning to rank based on subsequences}

\author{Basura Fernando{\thanks{The work was conducted at KU Leuven, ESAT-PSI.}}\\
ACRV, The Australian National University\\
\and
Efstratios Gavves\footnotemark[1]\\
QUVA Lab, University of Amsterdam, Netherlands\\
\and
Damien Muselet\\
Jean Monnet University, St-Etienne, France\\
\and
Tinne Tuytelaars\\
KU Leuven, ESAT-PSI, iMinds, Belgium\\
}

\maketitle

\begin{abstract}
We present a supervised learning to rank algorithm that effectively orders images by exploiting the structure in image sequences. Most often in the supervised learning to rank literature, ranking is approached either by analyzing pairs of images or by optimizing a list-wise surrogate loss function on full sequences. In this work we propose MidRank, which learns from moderately sized sub-sequences instead. These sub-sequences contain useful structural ranking information that leads to better learnability during training and better generalization during testing. By exploiting sub-sequences, the proposed MidRank improves ranking accuracy considerably on an extensive array of image ranking applications and datasets.
\end{abstract}

\section{Introduction}
\label{sec:intro}
 \begin{figure}[t]
\centering
\includegraphics[width=0.8\linewidth]{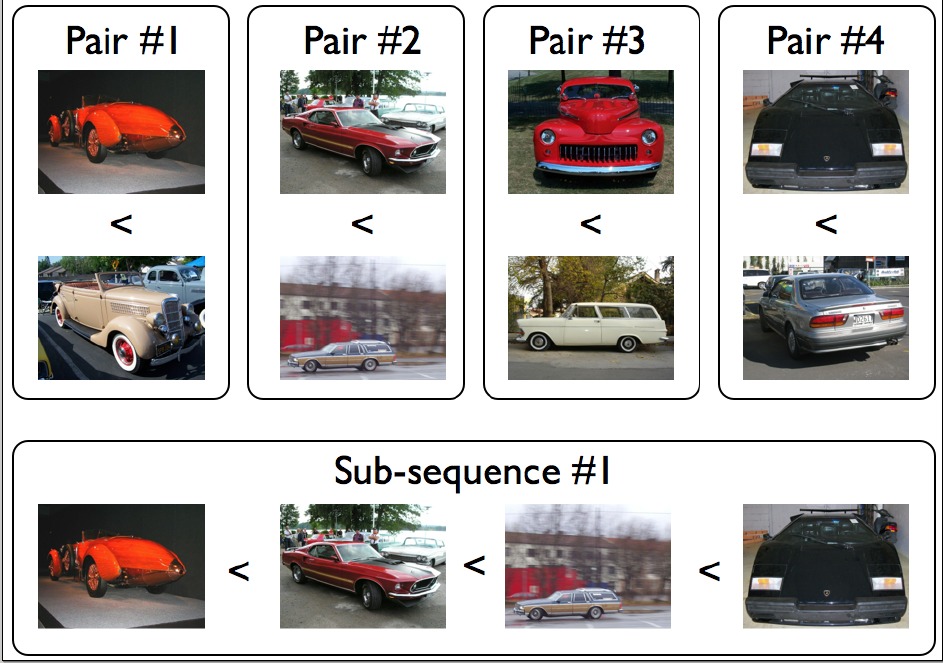}\\
\caption{Imagine you want to learn a ranking function from the example pairs in the top row. \emph{What could be the ranking criterion implied by the data?} Ranking \emph{more to less sporty cars} could be a criterion. Ranking \emph{more to less colorful} cars could be another. Given an example sequence, however, as in the bottom row, it becomes clear that the \emph{chronological} order is the most plausible ranking criterion. We advocate the use of such (sub-) sequences, instead of just pairs of images or long sequences, for supervised learning of rankers.}
\label{fig:intro}
\end{figure}

The objective of supervised learning-to-rank is to learn from training sequences a method that correctly orders unknown test sequences. This topic has been widely studied over the last years. Some applications include video analysis~\cite{Sun2014}, person re-identification~\cite{Wang2014a}, zero-shot recognition ~\cite{Parikh2011}, active learning~\cite{Liang2014}, dimensionality reduction~\cite{Deng2014}, 3D feature analysis~\cite{Tuzel2014}, binary code learning~\cite{Lin2014}, learning from privileged information~\cite{Sharmanska2013}, interestingness prediction~\cite{Fu2014} and action recognition using rank pooling~\cite{Fernando2015a}. In particular, we focus on image re-ranking. Image re-ranking is useful in modern image search tools to facilitate user specific interests by re-ordering images based on some user specific criteria. In this context, we re-order the top-k retrieved images for example from an image search~\cite{Fernando2013a,arandjelovic2012three}  based on some criteria such as interestingness~\cite{Gygli2013} or chronology~\cite{Lee2013,Fernando2015CVIU,Konstantinos2013a,Fernando2014b}. In this paper, we propose a new, efficient and accurate supervised learning-to-rank method that uses the information in image subsequences effectively for image re-ranking.

Most learning-to-rank methods rely on pair-wise cues and constraints. However, pairs of ranked images provide rather weak, and often ambiguous, constraints, as illustrated in Fig.~\ref{fig:intro}. Especially when complex and structured data is involved, considering only pairs during training may confuse the learning of rankers as shown in~\cite{Cao2007,Liu2009,Yisong7007}. Learning-to-rank is in fact a prediction task on lists of data/images. Treatment of pairs of images as independent and identically distributed random variables during training is not ideal~\cite{Cao2007}. It is, therefore, better, to consider longer subsequences within a sequence, which contain more information than pair of elements, see Fig.~\ref{fig:intro}.

To exploit the structure in long sequences, list-wise methods~\cite{Cao2007,Michael2008,Wu2010,xia2008listwise,Xu2007,Yisong7007}  
optimize for ranking losses defined over sequences. Although such an approach can exploit the structure in sequences, working on long sequences introduces a new problem. More specifically, as the number of wrong permutations grows exponentially with respect to the sequence length, list-wise methods often end up with a more difficult learning problem~\cite{Liu2009}. Hence, list-wise methods often lead to over-fitting, as we also observe in our experimental evaluations.

To overcome the above limitations, we propose to use \emph{subsequences} to learn-to-rank. On one hand, the increased length of subsequences brings more information and less uncertainty than pairs. On the other hand, compared to full sequences learning on subsequences allows more regularity and better learnability. 

Note that the training subsequences can be generated without any additional labeling cost: if the training set consists of sequences, we sub-sample; if the training set consists of pairs, we build subsequences by exploiting the transitivity property. We argue that every subsequence of any length sampled from a correctly ordered sequence also has the correct ordering i.e. \emph{all subsequences of a correctly ordered sequence are also correctly ordered}. We exploit this property as follows. Given the training subsequences, we learn rankers that minimize the \emph{zero-one loss} per subsequence length (or scale). During testing, using the above property, we evaluate all ranking functions of different lengths over the full test sequences using convolution. Then, to obtain the final ordering, we fuse the ranking results of different rankers.

Our major contributions are threefold. First, we propose a method, \emph{MidRank}, that exploits subsequences to improve ranking both quantitatively and qualitatively. Second,  we present a novel difference based vector representation that exploits the total ordering of subsequences. This representation, which we will refer to as \emph{stacked difference vectors}, is discriminative and results in learning accurate rankers. Third, we introduce an accurate and efficient polynomial time testing algorithm for the NP-hard~\cite{Schiavinotto2004} linear ordering problem to re-rank images in moderately sized sequences. 

We evaluate our method on three different applications: ranking images of famous people according to relative visual attributes~\cite{Parikh2011}, ranking images according to how interesting they are~\cite{Fu2014} and ranking car images according to the chronology~\cite{Lee2013}. Given an image search result obtained from an image search engine, we can use our method to re-rank images in a page to satisfy user specific criteria such as interestingness or chronology. Results show a consistent and significant accuracy improvement for an extensive palette of ranking criteria. This paper extends our the conference paper~\cite{Fernando2015}.

\section{Related work}
\label{sec:related}
Supervised learning-to-rank algorithms are categorized as point-wise, pair-wise and list-wise methods. Point-wise methods~\cite{Cooper1992}, which process each element in the sequence individually, are easy to train but prone to over-fitting. Pair-wise methods~\cite{herbrich00ordinal, Joachims2006, Sculley09} compute the differences between two input elements at a time and learn a binary decision function that outputs whether one element precedes the other or vice-versa. These methods are restricted to pair-wise loss functions. Naturally, pair-wise methods do not explicitly exploit any structural information beyond what a pair of elements can yield. List-wise methods~\cite{Cao2007,Wu2010,Xu2007,Yisong7007,Michael2008,xia2008listwise}, on the other hand formulate a loss on whole lists, thus being able to optimize more relevant ranking measures like the NDCG or the Kendall-Tau.

We present \svmmid, which belongs to a fourth family of \lrank methods, that is positioned between pair-wise and list-wise methods. Similar to pair-wise methods, \svmmid uses pairwise relations but extends to more informative subsequences, by considering multiple pairs within a subsequence simultaneously. Similar to list-wise methods, \svmmid optimizes a list-wise ranking loss, but unlike most list-wise methods we use zero-one sequence loss. This is done at sub-sequences thus allowing to exploit the \emph{regularity} in them during learning.

In~\cite{Dokania2014} Dokania \etal propose to optimize average precision \emph{information retrieval loss} using point-wise and pair-wise feature representations. However, this method only focuses on information retrieval.

\svmmid is also different from existing methods that use multiple weak rankers, such as LambdaMART~\cite{Wu2010} and AdaRank~\cite{Xu2007}, which propose a linear combination of weak rankers, with iterative re-weighting of the training samples and rankers during training. In contrast, \svmmid learns multiple ranking functions, one for each subsequence length, therefore focusing more on the latent structure inside the subsequences.

\section{\svmmid}
\label{sec:ranking}
We start from a training set of ordered image sequences. Each sequence orders the images according to a predetermined criterion, \eg images ranging from the most to the least happy face or from the oldest to the most modern car. Our goal is to learn from data in a supervised manner a ranker, such that we can order \emph{a new} list of \emph{unseen} images according to the same criterion.\\

\noindent\textbf{Basic notations.} Our training set is composed of $N$ ordered image sequences, $D=\{\mb{X}^i, \mb{Y}^i, \ell^i\}, i=1, \dots,N$. $\mb{X}^i$ stands for an image sequence $[\mb{x}^i_1, \mb{x}^i_2, \dots, \mb{x}^i_{\ell^i}]$ containing $\ell^i$ images, where $\ell^i$ can vary for different sequences $\mb{X}^i$. $\mb{Y}^i$ is a permutation vector $\mb{Y}^i=[\pi(1), ..., \pi(\ell^i)]$, and represents that the correct order of the images in the sequence is $\mb{x}^i_{\pi(1)} \succ \mb{x}^i_{\pi(2)} \succ \dots \succ \mb{x}^i_{\pi(\ell^i)}$. Henceforth, whenever we speak of a sequence $\mb{X}^i$, \emph{we imply that it is unordered}, and when we speak of \emph{an ordered sequence, we imply a tuple} $\{\mb{X}^i, \mb{Y}^i, \ell^i\}$. To reduce notation clutter, whenever it is clear from the context we drop the superscript $i$ referring to the $i$-th sequence.

\subsection{Ranking sequences}
\label{sec:inference}

Assume a new list of previously unseen images, $\mb{X}'=[\mb{x}_1', ..., \mb{x}_{\ell'}']$.
We define a ranking score function $z(\mb{X'}, \mb{Y'})$, which should return the highest score for the correct order of images $\mb{Y'}^*$. Given an appropriate loss function $\delta(\cdot, \cdot)$ our learning objective is
\begin{equation}
\argmin_{\vartheta} \delta(\mb{Y'}^*,\widehat{\mb{Y}}'),
\label{inference}
\end{equation}
\begin{equation}
\widehat{\mb{Y}}^{'}= \argmax_{ \mb{Y}^{'} } z(\mb{X}^{'}, \mb{Y}^{'}; \vartheta), 
\label{score}
\end{equation}
where $\widehat{\mb{Y}}^{'}$ is the highest scoring order for $\mb{X}'$ and $\vartheta$ are the parameters for our ranking score function.

The score function $z(\mb{X'}, \mb{Y'}; \vartheta)$ should be applicable for any length $\ell'$ that a new sequence $\mb{X'}$ might have.
To this end we decompose the sequence $\mb{X'}$ into a set of subsequences of a particular length, $\lambda$.
We only consider consecutive subsequences, \ie subsequences of the form $\mb{X_j}' = [\mb{x'}_{j:j+\lambda}]$.
The following proposition holds for any $\lambda \in [2...\ell']$:
\begin{prop}
A sequence of length $\ell$ is correctly ordered if and only if all of its $\ell-\lambda+1$ consecutive subsequences of length $\lambda$ are correctly ordered.
\label{myprop}
\end{prop}
This proposition follows easily from the transitivity property of inequalities.
We have, therefore, transformed our goal from ranking an unconstrained sequence of images, to ranking images inside each of the constrained, length-specific subsequences. To get to the final ranking of the original sequence we need to combine the rankings from the subsequences. Based on proposition~\ref{myprop}, we define the ranking inference as
\begin{equation}
\widehat{\mb{Y}}'= \argmax_{\mb{Y}'} \sum_{j=1}^{\ell-\lambda+1} z(\mb{X}^{'}_j, \mb{Y}^{'}_j; \vartheta).
\label{score-decomposed}
\end{equation}
with $\mb{Y}^{'}_j = [\pi^{'}(j)... \pi^{'}(j+\lambda-1)]$ and $z(\cdot)$ the ranking score function for fixed-length subsequences. The simplest choice for $z(\cdot)$ would be a linear classifier, \ie $z(\mb{X}^{'}, \mb{Y}^{'}; \vartheta) = \vartheta^T \psi(\mb{X}^{'}, \mb{Y}^{'})$, where $\psi(\mb{X}^i, \mb{Y}^i)$ is a feature function that we will revisit in the next subsection. However, since eq.~\eqref{score-decomposed} sums the ranking scores for all subsequences together, a non-linear feature mapping is to be preferred, otherwise the effect of different subsequences will be cancelled out. In practice, we use $z(\mb{X}^{'}, \mb{Y}^{'}; \vartheta) = sign (\vartheta^T \psi(\mb{X}^{'}, \mb{Y}^{'})) . |\vartheta^T \psi(\mb{X}^{'}, \mb{Y}^{'})|^{\frac{1}{2}}$. It is worth noting that the ranking inference of eq.~\eqref{score-decomposed} is a generalization of the inference in pairwise ranking methods, like RankSVM~\cite{Joachims2006}, where $\lambda=2$. Moreover, eq.~\eqref{score-decomposed} is similar in spirit to convolutional neural network models~\cite{NIPS2012_4824}, which decompose the input of unconstrained size to a series of overlapping operations.

\subsection{Training $\lambda$-subsequence rankers}

Having decomposed an unconstrained ranking problem into a combination of constrained, length-specific subsequence ranking problems, we need a learning algorithm for optimizing $\vartheta$. Considering prop.~\eqref{myprop} and eq.~\eqref{score-decomposed} we train the parameters $\vartheta$ for a given subsequence length, as follows
\begin{align}
\argmin_{\vartheta} & \frac{\mu}{2} \|\vartheta\|^2+ \sum_{i,j} \mathcal{L}^i_{(\lambda)j} \nonumber\\
\mathcal{L}^i_{(\lambda)j} =& \max\{0, 1-\delta(\mb{Y}^i_j, \mb{Y}^{i\ast}_j) \; \vartheta^T  \cdot
\psi(\mb{X}^i_{j}, \mb{Y}^i_j)\}.
\label{opt-func}
\end{align}
The loss function $\mathcal{L}^i_{(\lambda)j}$ measures the loss of the $j$-th subsequence of length $\lambda$ of the $i$-th original sequence.
For discriminative learning in eq.~\eqref{opt-func} we need both positive and negative subsequences. During training we sample subsequences of standardized lengths. Although we can mine positive subsequences of all possible lengths, in practice we focus on subsequences up to length 7-10. To generate negative subsequences during training, we scramble the correct order randomly sampled subsequences. For each positive subsequence of length $\lambda$, we can generate theoretically up to $\lambda!-1$ negatives. However, this would create a heavily imbalanced dataset, which might influence the generalization of the learnt rankers~\cite{akata2014}. Furthermore, keeping all possible negative subsequences would have a severe impact on memory. Instead, we generate as many negative subsequences as our positives. For the optimization of eq.~\eqref{opt-func} we use the stochastic dual coordinate ascent (SDCA) method~\cite{Shwartz2011}, which can handle imbalanced or very large training problems~\cite{Shwartz2011}. In practice, when adding more negatives we did not witness any significant differences.

The $\delta(\cdot)$ function operates as a weight coefficient. The optimization uses indirectly the ranking disagreements to emphasize the wrong classifications proportionally to the magnitude of the $\delta(\cdot)$ disagreement value. At the same time, the optimization is expressed as a zero-one loss based classification using hinge loss. This allows to maximize the margin between the correct and the incorrect subsequences of specific length.

As we are interested in obtaining the optimal ranking, we could implement $\delta(\cdot)$ using any ranking metric, such as sequence accuracy, Kendall-Tau or the NDCG. From the above metrics the sequence accuracy, $\delta(\mb{Y}, \mb{Y'}^*)= 2 [\mb{Y}=\mb{Y'}^*]-1$ is the strictest one, where $[\cdot]$ is the Iverson's bracket. List-wise methods usually employ relatively relaxed ranking metrics, \eg based on the NDCG or Kendall-Tau measure. This is mostly because for longer, unconstrained sequences on which they operate directly, the zero-one loss is too restrictive. In our case, the advantage is we have standardized, relatively small, and length-specific subsequences, on which the zero-one loss can be easily applied. With zero-one loss we observe better discrimination of correct subsequences from incorrect ones and, therefore, better generalization. To this end in our implementation we use the zero-one loss, although other measures can also be easily introduced. 

\subsection{Ranking-friendly feature maps}
\label{sec:representations}

To learn accurate rankers we need discriminative feature representations $\psi(\mb{X}^i_j, \mb{Y}^i_j)$. We discuss three different representations for $\mb{X}_{(\lambda)}$.\\

\noindent\emph{Mean pairwise difference representation.} Herbrich \etal~\cite{herbrich00ordinal} eloquently showed that \emph{the difference of vector} representation yields accurate results for learning a pairwise ranking function.

For this representation we have $\psi(\mb{X}, \mb{Y})=\frac{1}{|\{(i,j)|\mb{x}_{\pi(i)} \succ \mb{x}_{\pi(j)} \}|} \sum_{\forall \{i,j| \mb{x}_{\pi(i)} \succ \mb{x}_{\pi(j)} \} } (\mb{x}_{\pi(i)} - \mb{x}_{\pi(j)})$. The mean pairwise difference representation is perhaps the most popular choice in the learning-to-rank literature~\cite{herbrich00ordinal, Smola2004, Joachims2006, Parikh2011, Lin2014,Dokania2014}. In the specific case of $\lambda=2$ we end up with the standard rank SVM~\cite{Joachims2006} and the SVR~\cite{Smola2004} learning objectives.\\

\noindent\emph{Stacked representation.} Standard pairwise max margin rankers make the simplifying assumption that a sequence is equivalent to a collection of pairwise inequalities. To exploit the structural information beyond pairs, we propose to use the stacked representation as $\psi(\mb{X}, \mb{Y})=[\mb{x}^T_{\pi(1)} , \ldots,  \mb{x}^T_{\pi(\ell)}]^T$ for subsequences.\\

An interesting property of stacked representations comes from the field of combinatorial geometry~\cite{birkhoff}. From combinatorial geometry we know that all permutations of a vector are vertices of a \textit{Birkhoff polytope}~\cite{birkhoff}. As the Birkhoff polytope is provably convex, there exists at least one hyperplane that separates each vertex/permutation from all others. Hence, there exists also at least one hyperplane to separate the optimally permutated sequence $\mb{X}^*_l$ from all others, \ie, we have a linearly separable problem. Of course this linear separability applies only for the different permutations of a particular list of elements $\mb{X}^i_l$. Thus is not guaranteed that all correctly permuted $\mb{X}^{i*}_l, \forall i$ training sequences will be linearly separable from all incorrectly permuted ones. However, this property ensures that from all possible permutations of a sequence, the correct one can always be linearly separated from the incorrect ones.

The same advantage of better separability between different orderings of the same sequence could be obtained by nonlinear kernels. Such kernels, however, are too expensive to apply on many realistic scenarios, when thousands of sequences are considered at a time. Furthermore, the design of such kernels is application dependent, thus making them less general.\\

\noindent\emph{Stacked difference representation.} Inspired from \cite{herbrich00ordinal} and the stacked representations, we can also represent a sequence of images as $\psi(\mb{X}, \mb{Y})=[(\mb{x}_{\pi(1)}-\mb{x}_{\pi(2)})^T , \ldots, (\mb{x}_{\pi(\lambda-1)}-\mb{x}_{\pi(\lambda)})^T]^T$. Similar to mean pairwise difference representations, they model only the difference between neighboring elements in a rank, thus being invariant to the absolute magnitudes of the elements in $\mathbf{X_\ell}$. Furthermore it is easy to show that stacked difference representations maintain total order structure (proof in supplementary material\footnote{\url{users.cecs.anu.edu.au/~basura/ICCV15_sup.pdf}}).
As a result, if there is some latent structure in the subsequence explaining why a particular order is correct, the stacked difference representation will capture it, to the extent of the feature's capacity. 

\subsection{Multi-length \svmmid}
\label{sec.mvv}

So far we focused on subsequences of fixed length $\lambda$.
A natural extension is to consider multiple subsequence lengths, as different lengths are likely to capture different aspects of the example sequences and subsequences. To train a multi-length ranker we simply consider each length in eq.~\eqref{opt-func} separately, namely $\lambda=2, 3, 4, \dots, L$. To infer the final ranking we need to fuse the output from the different length rankers. To this end we propose a  weighted majority voting scheme.

For each test instance $\mb{X}'$ we obtain a ranking per $\lambda$ and the respective ranking score from $z(\cdot)$. As a result we have rankings for each of the $L-1$ rankers. Then, each image in the test sequence gets a vote for its particular position as returned from each ranker. Also, each image gets a voting score that is proportional to the ranking score from $z(\cdot)$, and, therefore, the confidence of the ranker for placing that image to the particular position. We weight the image position with the voting scores and compute the weighted votes for all images for all positions. Then, we decide the final position of each image starting rank 1, selecting the image with the highest weighted vote at rank 1. Then we eliminate this image from the subsequence comparisons and continue iteratively, until there are no images to be ranked.

\section{Efficient inference}
\label{sec.theory.inference.main}
\begin{figure}[t]
\centering
\includegraphics[width=0.99\linewidth]{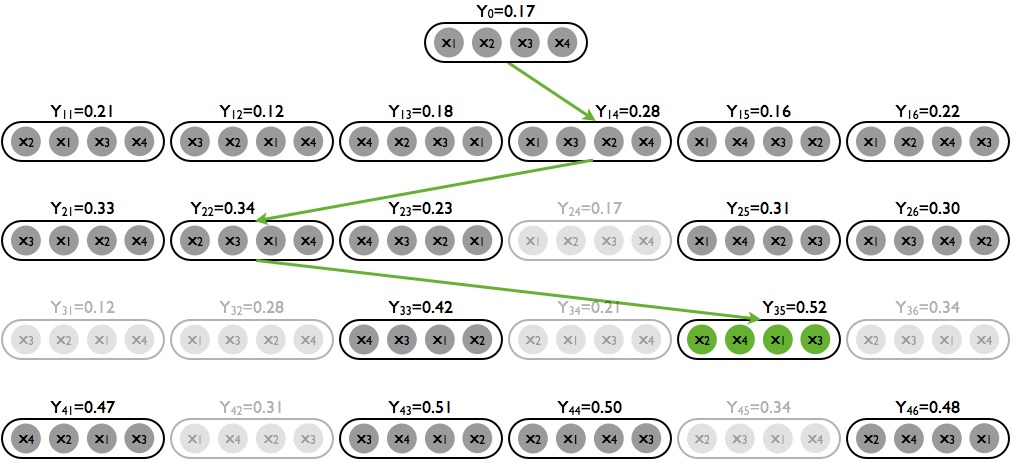}
\caption{Illustration of the proposed greedy inference algorithm. Starting from 
the root, we visit all child permutations where only a single pair is switched. 
Then we select the child permutation with the maximum score. If this maximum 
score is greater than the parent score, then we expand the maximum child, 
otherwise we stop. In the process we do not revisit the permutations that have 
already been visited, see faded nodes. We limit the search procedure to a 
maximum of $l$ depths of the tree.}
\label{fig:greedy}
\end{figure}

Solving eq.~\eqref{score-decomposed} requires an explicit search over the space of all possible permutations in $\mb{Y}$, which amounts to $l!$. Hence, for a successful as well as practical inference we need to resolve this combinatorial issue. Inspired by random forests~\cite{Breiman2001} and the best-bin-first search strategies~\cite{Beis1997}, we propose the following greedy search algorithm. For a visual illustration of the algorithm we refer to Figure~\ref{fig:greedy}.

We start from an initial ranking solution $\widehat{\mb{Y}}'^{(0)}$ obtained from a less accurate, but straightforward ranking algorithm (\ie RankSVM). Given $\widehat{\mb{Y}}'^{(0)}$, we generate a set of permutations denoted by $\{\widehat{\mb{Y}}'^{(1)}\}$, such that the new permutations are obtained by only swapping a pair of elements of $\widehat{\mb{Y}}'^{(0)}$. From all permutations of $\{\widehat{\mb{Y}}'^{(1)}\}$, we compute the ranking scores using Eq.~\ref{score-decomposed} and pick the permutation with the maximum score denoted by $\widehat{\mb{Y}}'^{(1)}$. If this score is larger than the score of the parent permutation (i.e. $ \sum z(\mb{X}^{'}_j, \widehat{\mb{Y}}'^{(1)}_j; \vartheta, \lambda) >  \sum z(\mb{X}^{'}_j, \widehat{\mb{Y}}'^{(0)}_j; \vartheta, \lambda)$), we set the permutation $\widehat{\mb{Y}}'^{(1)}$ as the new parent and traverse the solution space recursively using the same strategy. Permutations that have already been visited are removed from any future searches. The procedure of traversing through the above strategy forms a tree of permutations (see Figure~\ref{fig:greedy}). We stop when \emph{i)} no other solution with a higher score can be found (score criterion) or \emph{ii)} when we reach the $l$-th level of the permutation tree (depth criterion).

At each level of the tree, we traverse a maximum of $\ell^i \cdot (\ell^i-1)/2$ nodes. Experimentally, we observe that after only a few levels we satisfy the score criterion, thus having an average complexity of $O({\ell^i}^2)$. When the depth criterion is satisfied, we have the worst case complexity of $O({\ell^i}^3)$. 

To further ensure that the search algorithm has not converged to a poor local optimum, we repeat the above procedure starting from different initial solutions. For maximum coverage of the solution space, we carefully select the new $\widehat{\mb{Y}}'^{(0)}$, such that $\widehat{\mb{Y}}'^{(0)}$ was not seen in the previous trees. As we also show in the experiments, the proposed search strategy allows for obtaining good results using very few trees. Inarguably, our efficient inference enables ranking with subsequences, as the exhaustive search is impractical for moderately sized sequences (8-15 elements long), and intractable for longer sequences. \\

\section{Experiments}
\label{sec:ranking-experiments}
\begin{table*}[t!]
\centering
 \scriptsize
\begin{tabular}{l c c c c c c c c c c c}
\toprule
 & \multicolumn{3}{c}{\textbf{Public figures}} & & \multicolumn{3}{c}{\textbf{Scene interestingness}} & & \multicolumn{3}{c}{\textbf{Car chronology}} \\
 \cmidrule{2-4} \cmidrule{6-8} \cmidrule{10-12}
\textit{Method}		& \textit{NDCG} & \textit{KT} & \textit{Pair Acc.} & & \textit{NDCG} & \textit{KT} & \textit{Pair Acc.} & & \textit{NDCG} & \textit{KT} & \textit{Pair Acc.} \\
\midrule
\textit{SVR}			& .882 & .349 &  65.7 & & .870 & .317 & 65.8 & & .910 & .399 & 69.9 \\
\textit{McRank}			& .921 & .540 &  76.9 & & .859 & .295 & 64.8 & & .921 & .477 & 70.6 \\

\textit{RankSVM} 		& .947 & .617 & 80.8 & & .870 & .317 & 65.8 & & .928 & .482 & 74.1 \\
\textit{Rel. Attributes}	& .947 & .616 & 80.8 & & .869 & .315 & 65.7 & & .927 & .479 & 73.9 \\
\textit{CRR}			& .945 & .612 &	80.6 & & .846 & .273 & 63.6 & & .912 & .394 & 69.7 \\

\textit{AdaRank}		& .836 & .154 & 57.7 & & .745 & -.077 & 46.1  & & .827 & .118 & 55.9 \\
\textit{LambdaMART}		& .855 & .207 & 60.4 & & .860 & .315  & 64.3  & & .935 & .409 & 70.6 \\
\textit{ListNET} 		& .866 & .314 & 65.7 & & .821 & .118  & 55.9  & & .872 & .291 & 64.5 \\
\textit{ListMLE}        	& .851 & .262 & 63.1 & & .862 & .282  & 64.1  & & .854 & .278 & 63.9 \\
\midrule
\textit{\svmmid} 	 	& \bt{.954} & \bt{.722} & \bt{84.7} & & \bt{.887} & \bt{.347} & \bt{67.4} & & \bt{.949} & \bt{.553} & \bt{76.9} \\
\bottomrule
\end{tabular}
\caption{Evaluating ranking methods on three datasets and applications: ranking public figures using relative attributes of~\cite{Parikh2011}, ranking scenes according to how interesting they look~\cite{Gygli2013} and ranking cars according to their manufacturing date~\cite{Lee2013}. For public figures we have sequences of 8 images because of the size of the dataset. For the scenes and the cars datasets we have sequences of 20 images. Similar trends were observed with sequences of 80 images. For all baselines we use the publicly available code and the recommended settings.}
\label{tab:allnumbers}
\end{table*}

We select three supervised \emph{image re-ranking applications} to compare \svmmid with other supervised learning-to-rank algorithms, namely, 
ranking public figures (section~\ref{sec.rank.exp.public}), ordering images based on interestingness (section~\ref{sec.rank.exp.inter}) and chronological ordering of car images (section~\ref{sec.rank.exp.cars}). Next, we analyze the properties and the efficiency of \svmmid under different parameterizations in section~\ref{sec.rank.exp.detailed}. 

\begin{figure*}[t]
\centering
\includegraphics[width=1.0\linewidth]{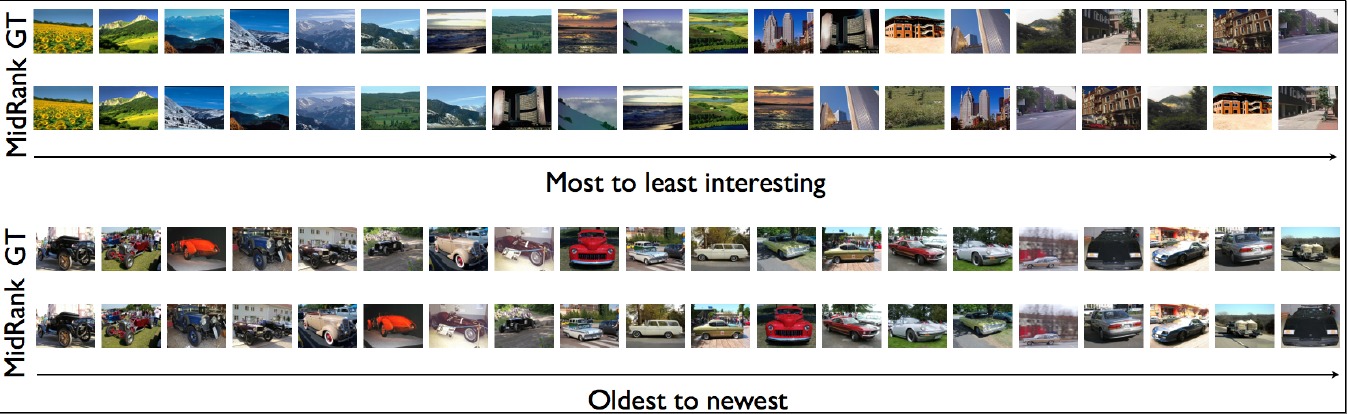}
\caption{Example of ordering images with \svmmid according to how interesting they look, or how old is the car they depict. Although both tasks seem rather difficult with the naked eye, \svmmid returns rankings very close to the ground truth. We include more visual results in the supplementary material.}
\label{fig:sample}
\end{figure*}

\subsection{Evaluation criteria \& implementation details} 
\label{sec:rank.exp.eval.criteria} We evaluate all methods with respect to the 
following ranking metrics. First, we use the \textit{normalized discounted 
cumulative gain} NDCG, commonly used to evaluate ranking 
algorithms~\cite{Liu2009}. The discounted cumulative gain at position $k$ is 
defined as $DCG@k=\sum_{i=1}^k\frac{2^{rel_i}-1}{\log_2(i+1)}$, where $rel_i$ is 
the relevance of the image at position $i$. To obtain the normalized DCG, the 
$DCG@k$ score is divided by the ideal DCG score. NDCG, whose range is $[0, 1]$, 
is strongly \textit{non-linear}. For example going from 0.940 to 0.950 
indicates a significant improvement.  

We also use the \textit{Kendall-Tau}, which captures better how close we are to the perfect ranking. The \textit{Kendall-Tau} accuracy is defined as $KT=\frac{l^{+}-l^{-}}{0.5l(l-1)}$, where $l^{+}, l^{-}$ stand for the number of all pairs in the sequence that are correctly and incorrectly ordered respectively, and $l=l^{+}+l^{-}$. \textit{Kendall-Tau} varies between $-1$ and $+1$ where a perfect ranker will have a score of $+1$. For completeness we also use pairwise accuracy as an evaluation criterion in which we count the percentage of correctly ranked pairs of elements in all sequences. 

We compare our \svmmid with point-wise methods such as SVR~\cite{Smola2004}, McRank~\cite{li2007mcrank}, pair-wise methods such as RankSVM~\cite{Joachims2006}, Relative Attributes~\cite{Parikh2011} and CRR~\cite{Sculley09}. We also compare with list-wise methods such as AdaRank~\cite{Xu2007}, LambdaMART~\cite{Wu2010}, 
ListNET~\cite{Cao2007} and ListMLE~\cite{xia2008listwise}. For all methods we use the publically available code as provided by the authors, the same features and recommended settings for fine-tuning. All these baselines and \svmmid take the same set of training sequences as input. There is no overlap between elements of train and test sequences. All training sequences are sampled from the training set of each dataset and testing sequences are sampled from the testing set. We make these train and test sequences along with the data publicly available. We evaluate all methods on a large number of 20,000 test sequences. We experimentally found that \emph{the standard deviations are quite small for all methods}. 

For \svmmid we pick the values for any hyper-parameters (\eg the cost parameter in eq.~\eqref{opt-func}) after cross-validation. We investigate \svmmid rankers of length $3-8$ and merge the results with the majority weighted voting, as described in Sect.~\ref{sec.mvv}. For the efficient inference, we initialize the parent node with the solution obtained from the pair-wise RankSVM~\cite{Joachims2006}. We also tried various ranking score normalizations between the different length rankers, but we found experimentally that results did not improve significantly. Consequently, we opted for directly using the unnormalized ranking scores from different length rankers. In all cases the feature vectors are L2-normalized.

\subsection{Ranking Public Figures}
\label{sec.rank.exp.public}

First we evaluate \svmmid on ranking public figure images with respect to 
relative visual attributes~\cite{Parikh2011}, using the features and the 
train/test splits provided by~\cite{Parikh2011}. The dataset consists of images 
from eight public figures and eleven visual attributes of theirs, such as 
\textit{big lips}, \textit{white} and \textit{chubby}. Our goal is to learn 
rankers for these attributes. Since there 8 public figures,  
we report results on the longest possible test sequence size composed of 8 
images. For each of the 11 attributes we sample 10,000 train sequences of length 8 and 
20,000 test sequences, totaling to 220,000 test sequences for all attributes. We use the standard GIST features provided with the dataset.
The results are reported by taking the average over all eleven attributes
and over all test sequences. See results in Table~\ref{tab:allnumbers}.

We observe that \svmmid improves the accuracy of the ranking significantly for 
all the evaluation criteria. For Kendall-Tau, \svmmid brings a +10.5\% absolute 
improvement. 
It is worth mentioning that for this dataset the best individual \svmmid function 
was of length 7, which in isolation scored a 0.684 Kendall-Tau accuracy.





\subsection{Ranking Interestingness}
\label{sec.rank.exp.inter}

Next, we evaluate \svmmid on ranking images according to how interesting
they are. We consider train and test sequences of size 20. It is not possible to consider much longer sequences as the annotation pool for interestingness is limited in this dataset. In practice even for humans rating more than 20 images based on interestingness would be difficult. 
We use the scene categories dataset from~\cite{Oliva2001}, whose 
images were later annotated with respect to interestingness by~\cite{Gygli2013}. 
We extract GIST~\cite{Oliva2001} features and construct 10,000 train sequences 
and 20,000 test sequences. Note that this is a difficult task, as interestingness is a subjective 
criterion which can be attributed to many different factors within an image. See results in Table~\ref{tab:allnumbers}.

We observe that also for this dataset \svmmid has a significantly better 
accuracy than the competitor methods for all the evaluation criteria. For visual results obtained from our method, see Fig.~\ref{fig:sample} (random example). As we can see, \svmmid returns a 
rank which visually is very close to the ground truth.



\subsection{Chronological ordering of cars}
\label{sec.rank.exp.cars}

As a final application, we consider the task of re-ranking images 
chronologically. We use the car dataset of~\cite{Lee2013}. The chronology of 
the cars is in the range \textit{1920, 1921, ..., 1999}. As image representation we use 64-Gaussian 
Fisher vectors~\cite{Perronnin2010} computed on dense SIFT features, after 
being reduced to 64 dimensions with PCA. To control the dimensionality we also 
reduce the Fisher vectors to 1000 dimensions using PCA again. Similar to the
previous experiment, we generate 10,000 training sequences and 20,000 test 
sequences of length 20. See results in Table~\ref{tab:allnumbers}.

Again, \svmmid obtains significantly better 
results than the competitor methods for all the evaluation criteria and 
especially for the Kendall-Tau accuracy (+7.1\%). We show some visual 
results in Fig.~\ref{fig:sample}. We also experimented with training sequence lengths of 5, 10, 15, 20 and testing sequence lengths of 5, 10, 20, 80. Due to brevity and space, we report on test sequences of length 20 only (which seems a more practical scenario in image search applications). However, in all these cases \svmmid outperforms all other methods. Note that despite the 
uncanny resemblance between the cars, especially the older ones, \svmmid returns 
a ranking very close to the true one.

\subsection{Detailed analysis of \svmmid}
\label{sec.rank.exp.detailed}

\noindent\textbf{Effect of subsequence length on \svmmid.}
\begin{figure}[t]
\centering
\subfloat[Effect of subsequence size]
{
\includegraphics[width=0.49\linewidth]{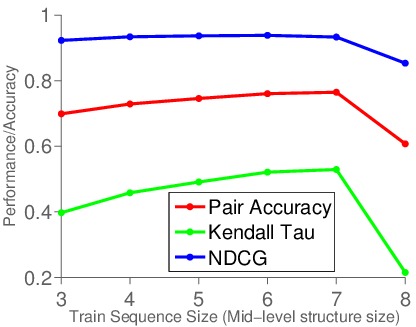}
}
\subfloat[Evaluating rank fusion methods]
{
\includegraphics[width=0.49\linewidth]{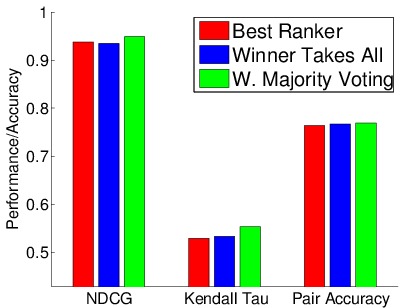}
}
\caption{(a) Evaluation of different individual \svmmid ranking functions of 
lengths 3 to 8 on the chronological ordering of cars task. We show how pair 
accuracy, Kendall Tau and NDCG vary over test sequences of size 20. (b) 
Comparison of several rank fusion strategies. Weighted majority voting method is 
the most effective strategy for combining rankers.}
\label{fig:effectofmid}
\end{figure}
Next, we evaluate the relation between the \svmmid accuracy and the training 
subsequence sizes. We use sequences of size 20 for training and testing 
generated from the car dataset. We evaluate different \svmmid ranking 
functions of size 3 up to 8. We plot the results in 
Fig.~\ref{fig:effectofmid}(a). 

For all ranking measures the best \svmmid is of size 7. Interestingly, the ranking performance gradually increases up to a point as the training subsequence size increases. This indicates that \svmmid ranking 
models trained on moderately sized subsequences perform better than very small or very large ones. Small \svmmid models are easy to train (small training errors), but solve a relatively easy ranking sub-problem. In contrast, large \svmmid models are more difficult to train (larger training errors). Our experiments suggest that moderately sized subsequences are the most suitable for \svmmid. It is also interesting to see that all three ranking measures used are consistent (--see Fig.~\ref{fig:effectofmid}(a)). However, the sensitivity of Kendall Tau seems to be larger than the other two ranking criteria (NDCG and pair-accuracy).

\begin{figure}[t]
\centering
\subfloat[I][I]
{
\includegraphics[width=0.49\linewidth]{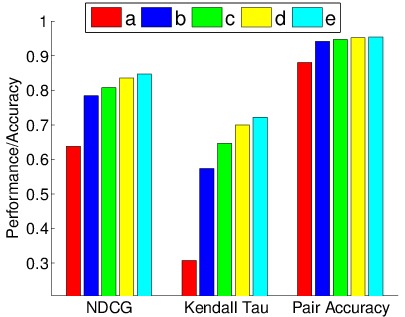}
}
\subfloat[II][II]
{
\includegraphics[width=0.49\linewidth]{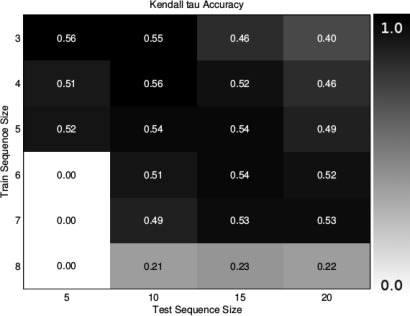}
}
\caption{(I) Comparing different representations for mid-level sequences detailed in~\ref{sec:representations} on the public figures dataset, also considering two more choices from the literature. : \textbf{(a)} max pooling of difference vectors as in~\cite{BoureauPL10}, \textbf{(b)} 
Mean pairwise difference representations, \textbf{(c)} the full stacked difference vector representation between all elements $i, j$ in a sequence, \textbf{(d)} the stacked representation, and \textbf{(e)} the stacked difference vector representation. The stacked difference vectors outperform all 
other alternatives. (II) How KT-accuracy changes when varying the train and test sequence lengths (cars dataset).}


\label{fig:stacked}
\end{figure}

\noindent\textbf{Evaluating subsequence representations.}
In this experiment we evaluate the effectiveness of the stacked difference 
representation introduced in section~\ref{sec:representations} 
compared to other alternatives see Fig.~\ref{fig:stacked} (left). The max-pooling or 
the mean pooling of difference vectors of a subsequence hinders useful ranking 
information, such as subtle variations between neighboring elements. Full stacked 
difference vectors (option (c)) does not perform as well as other stacked 
versions (d) and (e), probably due to the curse of dimensionality. Note that we also evaluated the mean representations on longer subsequences (20 images per sequence) and obtained 0.05 points lower in KT than the proposed stacked representations. This shows the best results are obtained with our stacked difference representation. 


\noindent\textbf{Evaluating efficient inference vs exhaustive inference.}
In this section we compare our efficient inference strategy with the exhaustive 
inference. As explained earlier, the exhaustive strategy has a complexity of 
$O(\ell^i!)$, whereas the proposed efficient inference strategy has an average 
complexity of $O({\ell^i}^2)$ and a worst case complexity of 
$O({\ell^i}^3)$.

\begin{figure}[t]
\centering
\subfloat[Time]
{
\includegraphics[width=0.49\linewidth]{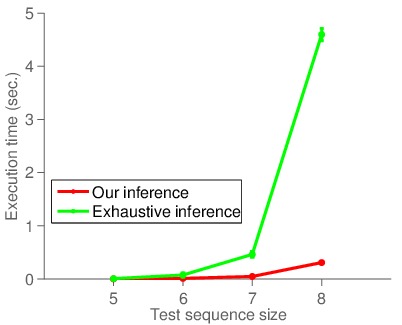}
}
\subfloat[Correctness]
{
\includegraphics[width=0.49\linewidth]{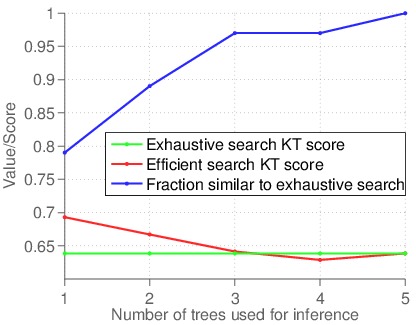}
}
\caption{(a) Comparison of execution time between the proposed efficient 
inference vs exhaustive search on public figures dataset. (b) Our efficient 
inference algorithm uses multiple trees. This figure shows how ranking performance (Kendall Tau) varies
with respect to the number of trees used. The blue plot shows the fraction of 
solutions (generated by the efficient algorithm) that agree with the solution 
obtained with exhaustive search.}
\label{fig:greedyTime}
\end{figure}

First, we plot how the execution time varies during inference for different 
test sequence sizes. We compare our inference method with the exhaustive search 
in Fig.~\ref{fig:greedyTime} (a). For moderately long test sequences, \eg up to size 8 in this experiment, our method is 50 times faster than exhaustive search. For longer sequences exhaustive inference is not even an option, as the number of possible combinations that need to be explored becomes very impractical, or even intractable for longer sequences.

Our inference method discovers the optimal order for a sequence of length 20 in 0.75 $\pm$ 0.1 seconds, and in practice sequences of up to 500 elements can be easily processed. Hence, \svmmid may easily be employed in the standard supervised image ranking and re-ranking scenarios, \eg improving the image search results based on user preferences.

In Fig.~\ref{fig:greedyTime}(b) we show significant improvement in execution time does not hurt the accuracy with respect to the exhaustive search. In this experiment we vary the number of trees used in our efficient algorithm and report the Kendal-Tau score and the percentage of sequences that agrees with the solution obtained with exhaustive search (blue line in Fig.~\ref{fig:greedyTime}(b)). Interestingly, using a single tree, we obtain a better Kendall-Tau score than the one obtained with the exhaustive search. We attribute this to some degree of over-fitting that might occur during learning. With 3 trees we obtain the same solution as exhaustive search for 97\% of the times, whereas with five trees we obtain exactly the same results as the exhaustive search.\\

\noindent\textbf{Evaluating Majority Voting Scheme.}
Last, we evaluate the effectiveness of different rank fusion strategies for 
\svmmid using the cars dataset. We compare the proposed weighted majority 
voting with \textit{winner-takes-all} strategy in which the ranker with the 
highest ranking score is used to define the final ordering. We also compare 
with the best individual ranker, where we use cross-validation to find the best 
ranker given a test sequence length.

As can be seen from Fig.~\ref{fig:effectofmid}(b), the weighted majority 
voting scheme works best. The results indicate that each ranker from 
different mid-level structure sizes exploits different types of structural 
information. Similar conclusions were derived for the other datasets.

\noindent\textbf{Evaluating on sequences of different lengths.}
In this experiment we evaluate how pair accuracy and KT vary for different train and test sequence sizes. We use the cars dataset for this experiment. From results reported in Fig.~\ref{fig:stacked} (right) we see that for smaller test sequences of 5 and 10 the best results are obtained using train subsequences of size 3 or 4. However, for larger test sequences of size 15 and 20 the best results are obtained for train subsequences of size 5, 6 and 7. Interestingly, the largest train subsequence size of 8 reports the worst results. These observations are valid for both pair accuracy as well as \textit{Kendall-Tau} performance.  

%
%


\section{Conclusion}
\label{sec:conclusion}
In this paper we present a supervised learning to rank method, \svmmid, that learns from sub-sequences. A novel \emph{stacked difference vectors} representation and an effective ranking algorithm that uses sub-sequences during the learning is presented. The proposed method obtains significant improvements over state-of-the-art pair-wise and list-wise ranking methods. Moreover, we show that by exploiting the structural information and the regularity in sub-sequences, \svmmid allows for a better learning of ranking functions on several image ordering tasks.
\\

\small{\textbf{Acknowledgement} Authors acknowledge the support from the Australian Research Council Centre of Excellence for Robotic Vision (project number CE140100016), the PARIS project (IWT-SBO-Nr.110067), iMinds project HiViz and ANR project SoLStiCe (ANR-13-BS02-0002-01).}


\end{document}